\documentclass[review]{elsarticle}

\usepackage{framed,multirow}
\usepackage{threeparttable}
\usepackage{booktabs,caption}
\usepackage{amssymb}
\usepackage{amsmath}
\usepackage{latexsym}
\usepackage{float}
\usepackage{url}
\usepackage{xcolor}
\usepackage{hyperref}
\usepackage{siunitx}
\usepackage{graphicx}
\usepackage{bm} 
\usepackage{multirow}

\hypersetup{pdfauthor=author, colorlinks=True}
\journal{Medical Image Analysis}

\biboptions{authoryear}

\begin{document}
\hypersetup{linkcolor=red,
            urlcolor=magenta,
            citecolor=blue}
\begin{frontmatter}

\title{Mutual Consistency Learning for Semi-supervised Medical Image Segmentation}

\author[1]{Yicheng Wu\corref{cor}}
\cortext[cor]{Corresponding Author}
\ead{yicheng.wu@monash.edu}
\author[2,3]{Zongyuan Ge}
\author[3]{Donghao Zhang}
\author[4]{Minfeng Xu}
\author[4]{Lei Zhang}
\author[5]{Yong Xia}
\author[1]{Jianfei Cai}
\address[1]{Department of Data Science \& AI, Faculty of Information Technology, \\Monash University, Melbourne, VIC 3800, Australia}
\address[2]{Monash-Airdoc Research, Monash University, Melbourne, VIC 3800, Australia}
\address[3]{Monash Medical AI, Monash eResearch Centre, Melbourne, VIC 3800, Australia}
\address[4]{DAMO Academy, Alibaba Group, Hangzhou, 311121, China}
\address[5]{National Engineering Laboratory for Integrated Aero-Space-Ground-Ocean Big Data Application Technology, School of Computer Science and Engineering, Northwestern Polytechnical University, Xi'an, 710072, China}
\begin{abstract}
In this paper, we propose a novel mutual consistency network (MC-Net+) to effectively exploit the unlabeled data for semi-supervised medical image segmentation. The MC-Net+ model is motivated by the observation that deep models trained with limited annotations are prone to output highly uncertain and easily mis-classified predictions in the ambiguous regions (\textit{e.g.,} adhesive edges or thin branches) for medical image segmentation.
Leveraging these challenging samples can make the semi-supervised segmentation model training more effective.
Therefore, our proposed MC-Net+ model consists of two new designs. 
First, the model contains one shared encoder and multiple slightly different decoders (\textit{i.e.,} using different up-sampling strategies). The statistical discrepancy of multiple decoders' outputs is computed to denote the model's uncertainty, which indicates the unlabeled hard regions.
Second, we apply a novel mutual consistency constraint between one decoder's probability output and other decoders' soft pseudo labels. In this way, we minimize the discrepancy of multiple outputs (\textit{i.e.,} the model uncertainty) during training and force the model to generate invariant results in such challenging regions, aiming at regularizing the model training.
We compared the segmentation results of our MC-Net+ model with five state-of-the-art semi-supervised approaches on three public medical datasets. Extension experiments with two standard semi-supervised settings demonstrate the superior performance of our model over other methods, which sets a new state of the art for semi-supervised medical image segmentation. Our code is released publicly at \url{https://github.com/ycwu1997/MC-Net}.
\end{abstract}


\end{frontmatter}

\section{Introduction}
Automatic medical image segmentation is a fundamental and critical step in constructing a powerful computer-aided diagnosis (CAD) system. According to the satisfied segmentation results, the morphological attributes of organs and tissues can be quantitatively analyzed to provide a useful basis for clinicians to diagnose diseases. At the same time, with an effective segmentation model, the localization of particular objects is significant for the early screening and precise assessment of relevant diseases \citep{background}.

Recent years have witnessed the remarkable progresses of deep learning for medical image segmentation. However, they still suffer from sub-optimal performance on many medical tasks. Their limited performance is mainly attributed to the over-fitting caused by inadequate training data, as most of medical image segmentation datasets are of a small scale. This is because acquiring adequate densely annotated medical data is extremely expensive. Manually annotating medical images (\textit{e.g.}, volumetric CT or MRI scans) at the pixel/voxel- level not only requires expertise and concentration but also is time-consuming. Therefore, exploiting unlabeled medical data like semi-supervised approaches has become considerably important to improve the performance of medical image segmentation models and attracted increasing research attention.

Existing semi-supervised methods can be roughly divided into two categories. The first approaches are the consistency-based models \citep{uamt,dtc,urpc} according to the smoothness assumption, \textit{i.e., small perturbations of an input should not produce the obvious deviations of corresponding outputs}  \citep{rampup}. The second category consists of several entropy-minimization methods \citep{pseudo,pseudolabeling,metapseudolabels}, which are based on the cluster assumption, \textit{i.e., the cluster of each class should be compact and thus of low entropy}. However, most of existing methods do not make full use of the learning difficulties \citep{curriculumlearning} of unlabeled data in semi-supervised tasks. Considering deep models can generate the segmentation results with the pixel/voxel- level uncertainties, we suggest leveraging such uncertainties to effectively exploit the unlabeled data, aiming at further improving the performance of semi-supervised medical image segmentation .
\begin{figure*}[htb]
\centering
\includegraphics[width=0.7\textwidth]{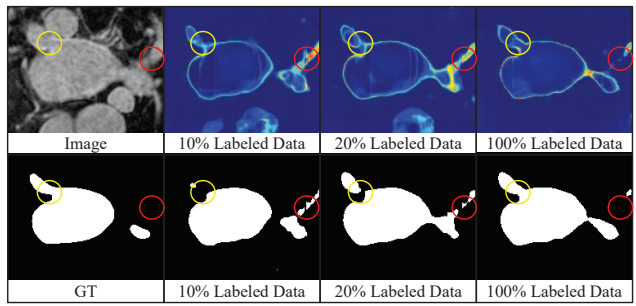}
\caption{\label{uncertainty}Three exemplar uncertainty maps and corresponding segmented results of a fully supervised V-Net model, trained with 10\%, 20\% and all labeled data on the LA dataset.}
\end{figure*}

Instead of following the curriculum learning \citep{curriculumlearning}, our main idea is to exploit the learning for unlabeled challenging regions to facilitate the model training. We further use Fig.~\ref{uncertainty} as an example to illustrate our motivation.
Specifically, Fig.~\ref{uncertainty} gives three uncertainty maps and segmentation results on the left artium (LA) dataset, which are obtained by three fully-supervised V-Net models, trained by 10\%, 20\% and all labeled data respectively. Each uncertainty map was obtained by the Monto-Carlo Dropout (MC-Dropout) method as \citep{uamt}.
Fig.~\ref{uncertainty} indicates two key observations: 
\textit{(1) The highly uncertain predictions are mainly located in some challenging regions (e.g., thin branch joints, indicated by the yellow and red circles in Fig.~\ref{uncertainty}).} Meanwhile, the regions without complex textures and varied appearances are more likely to be correctly segmented. In other words, trained with more labeled data, the V-Net model only refines the predictions of few hard areas;
\textit{(2) With the increase of labeled data for training, the model is prone to output less ambiguous results.} Thus, we hypothesize that the generalization ability of deep models should be highly related to the model uncertainty.
These observations motivate us to explore the model uncertainty to help the model generalize to these hard regions, which also aligns with a concurrent work in \citep{generalization}.

Therefore, in this paper, we propose a novel mutual consistency network (MC-Net+) for semi-supervised medical image segmentation, aiming to pay more attention to unlabeled challenging regions via the estimated model uncertainty. First, as Fig.~\ref{model} shows, our MC-Net+ model is composed of one shared encoder and multiple slightly different decoders. The statistical discrepancy of multiple decoders' outputs is used to represent the pixel/voxel- level uncertainty, indicating the hard regions. Second, we utilize a sharpening function to convert the probability outputs into soft pseudo labels. Then, we design a new mutual consistency training scheme, which enforces a consistency constraint between one decoder's probability output and other decoders' soft pseudo labels. In this way, we minimize the output discrepancy of multiple decoders during the model training and establish an `end-to-end' way to train our model, attempting to capture more useful features from unlabeled challenging regions.

Overall, our contributions of this paper are three-fold.
\begin{itemize}
\item We proposed the MC-Net+ model for semi-supervised segmentation, with the key idea that enforcing the model to generate consistent and low-entropy predictions in the hard regions can effectively exploit the unlabeled data and improve the semi-supervised image segmentation performance.
\item We designed a novel mutual consistency scheme to take advantage of both the consistency and entropy-minimization constraints for the model training, enabling the model to learn a generalized feature representation.
\item Extensive experiments demonstrate that the proposed MC-Net+ model outperforms five recent methods and sets a new state of the art (SOTA) for semi-supervised medical image segmentation.
\end{itemize}

The preliminary version of this work appeared in MICCAI 2021 \citep{mcnet}, which encourages the mutual consistency between two slightly different decoders. This paper substantially extends the conference version. The main extensions include: (1) embedding another decoder using a nearest interpolating operation into the original MC-Net model, which further increases the intra-model diversity; (2) conducting experiments on the extra Pancreas-CT and ACDC datasets to demonstrate the general effectiveness of our model on other semi-supervised medical image segmentation tasks; (3) implementing five recent approaches in the same environment and reporting their corresponding results for fair comparisons; (4) supplementing the hyper-parameter discussions; (5) adopting the original encoder-decoder architecture for testing, without introducing additional inference costs.

\section{Related Work}
\subsection{Semi-supervised Learning}
Semi-supervised learning (SSL) is widely studied in various computer vision tasks. For the consistency-based models, many data augmentation methods \citep{mixup, sharpening,ssnet} are used to generate different perturbed data. For example, \citet{cct} enforced several data augmentation operations to perturb the intermediate feature maps and constrained the model to output invariant segmentation maps. \citet{semanticda} utilized the semantic direction in the feature space to achieve semantic data augmentation and then applied consistency constraints for SSL. \citet{fixmatch} employed the consistency of training samples under weak and strong perturbations to facilitate the model training. Consistency at the model level is also discussed in the mean-teacher model via using an exponential moving average (EMA) operation \citep{meanteacher}. Meanwhile, the adversarial training \citep{vat,gan,you2022class} is used to enforce stronger consistency constraints for the model training.

Additionally, the entropy-minimization-based models can also boost semi-supervised learning. For instance, \citet{universalsss} proposed an entropy module to enable the model to generate low-entropy predictions in the unlabeled set. Furthermore, the pseudo label learning \citep{pseudo,cps} often employs a sharpening function or a fixed threshold to convert probability maps into pseudo labels. Then, supervised by pseudo labels, the model can learn to generate low-entropy results. For example, \citet{pseudolabeling} unitized the probability and uncertainty thresholds to select the most accurate pseudo labels for SSL. \citet{metapseudolabels} incorporated the meta-learning scheme into the pseudo label learning to improve performance.

It is nowadays widely recognized that both the consistency and entropy-minimization constraints can boost the feature discriminatory power of semi-supervised models. Therefore, in this paper, we employ both techniques in our MC-Net+ model for accurate semi-supervised medical image segmentation.

\subsection{Semi-supervised Medical Image Segmentation}
Several recent semi-supervised methods have been proposed for the medical image segmentation task. For example, \citet{uamt} proposed an uncertainty-aware mean-teacher model for semi-supervised left atrium segmentation. \citet{sassnet} further enforced the shape constraints via introducing the signed distance map (SDM) \citep{dtm} to improve the performance. Meanwhile, \citet{dtc} studied the relation between medical image segmentation and organ shape regression. They also investigated a semi-supervised model to achieve the multi-scale consistency for the gross target volume segmentation \citep{urpc}. Furthermore, \citet{cotraining,cotraining2} employed a multi-view co-training strategy to perform ensemble learning for 3D medical image segmentation. \citet{pairwise} utilized the attention mechanism to learn the pair-wise relation between labeled and unlabeled data to further relieve the over-fitting caused by limited labeled data.

Although these models have reported good results for semi-supervised medical image segmentation, they still neglect or underestimate the effects of the unlabeled challenging regions during the model training. In other words, we hypothesize that the performance of our task can be further improved via more effective modeling the challenging regions even without corresponding labels. Note that, we noticed that the CPS model \citep{cps} recently developed a cycled-consistency model similar to ours for semi-supervised image segmentation, but their model employs an identical model architecture with different initialization parameters and exploits different input noises to perturb input images. In contrast, our model is motivated by Fig.~\ref{uncertainty} and adopts a shared encoder with multiple slightly different decoders for training. Section \textit{6.1} further demonstrates that using different up-sampling strategies can lead to better segmentation results.

\subsection{Multi-task Learning}
Another research direction to improve the generalization of a deep model is through learning a cross-task feature representation or conducting an unsupervised pre-training \citep{you2021momentum}. The contrastive learning-based models \citep{mt8} can perform self-supervised training to mitigate the over-fitting of deep models. For example, \citet{you2022bootstrapping} employed global/local contrastive learning to extract more generalized features from unlabeled data and produced significant performance gains for semi-supervised medical image segmentation. Furthermore, some proxy or auxiliary tasks can be constructed to explicitly regularize the model training \citep{rubik,you2020unsupervised}. Specifically, the shape or boundary constraints can be used for the shape refinement to promote medical image segmentation \citep{dtm,mt7}. Some auxiliary losses (\textit{e.g.,} for image reconstruction) can also help the model extract more generalized and useful features \citep{mt5,mt6}. For instance, \citet{you2022simcvd} combined the knowledge distillation and multi-task learning to leverage the unlabeled data effectively, which achieved satisfied semi-supervised segmentation performance.

Compared to these successful methods, our proposed MC-Net+ model does not need to design specific auxiliary tasks and only considers the original segmentation task for the model training. On the other hand, our proposed method can be easily incorporated with those multi-task learning models to further boost semi-supervised medical image segmentation.
\subsection{Uncertainty Estimation}
The uncertainty analysis attracts much attention in the both fields of machine learning and computer vision \citep{unreview,assessing}. We not only expect the model to output correct results, but also hope to obtain the confidence of generated predictions. For example, the inherent aleatoric uncertainty is caused by the annotation noises and the epistemic uncertainty accounts for the discrepancy of deep models \citep{uncertainty}. In semi-supervised scenarios, we here only discuss the epistemic uncertainty, which can be reduced by giving more training data.

There are some existing methods to estimate the uncertainty. For example, \citet{uqwithunet} employed the variational U-Net \citep{varunet} to represent the model's uncertainty. The epistemic uncertainty can also be quantified via the model ensemble strategy \citep{modelensembling}, which computes the statistical discrepancy of different outputs by several individually trained models. However, this scheme would bring more computational costs. To address this, in bayesian modeling, the MC-Dropout method is proposed to approximate the model's uncertainty via a more flexible way \citep{mcdropout}. Specifically, the dropout operation samples multiple sub-models from the whole model. The statistical discrepancy of sub-models' outputs can be used to indicate the model's uncertainty. Thus, there is no need to train multiple models individually. In this paper, inspired by \citep{dass}, our model pre-defined multiple sub-models before the model training, which estimates the model's epistemic uncertainty in only one forward pass.

\section{Method}
\begin{figure*}[htb]
\centering
\includegraphics[width=1\textwidth]{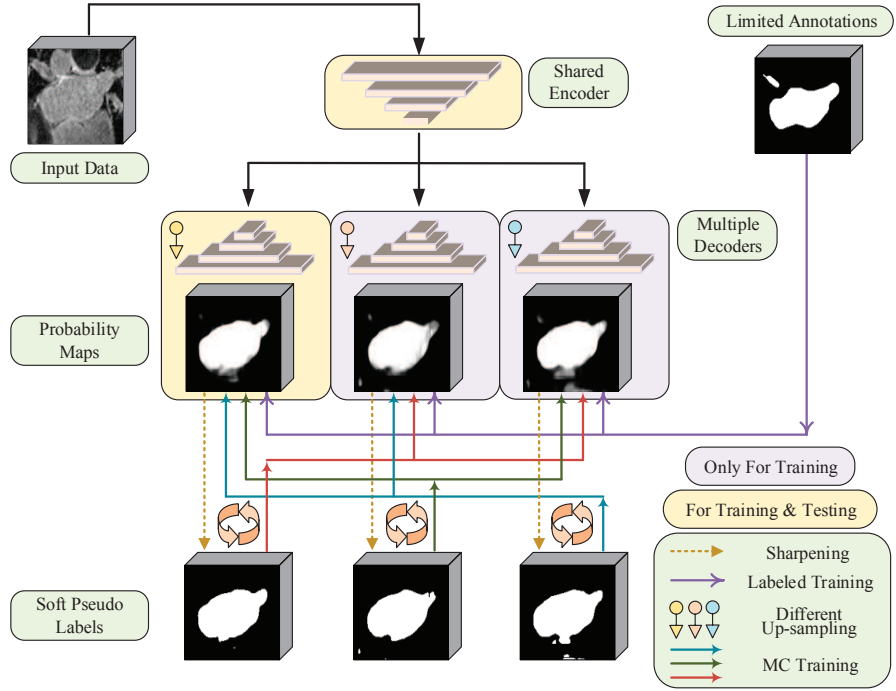}
\caption{\label{model}Diagram of our proposed MC-Net+ model, where the mutual consistency constraint is applied between one decoder's probability output and other decoders' soft pseudo labels. Note that, there are three slightly different decoders in this paper.}
\end{figure*}
Before introducing our model, we first define the semi-supervised segmentation problem with a set of notations. We use $x \in X$ to denote an input image and $p(y_{pred}|x; \theta)$ is the generated probability map of $x$, where $\theta$ denotes the parameters of a backbone $f_{\theta}$. Let $y_l\in Y_l$ denote the given segmentation annotations. The labeled and unlabeled sets are denoted as $\mathbb{D_L}$ = $\lbrace x_l^{i}, y_l^{i}|i=1, ..., N_l \rbrace$ and $\mathbb{D_U}$ = $\lbrace x_u^{i}|i=1, ..., N_u \rbrace$, respectively.
\subsection{Model Architecture}
The proposed MC-Net+ model attempts to exploit the unlabeled ambiguous regions for the model training, which can be indicated via the model's epistemic uncertainty. Essentially, the normal uncertainty estimation process can be defined as follows:
\begin{equation}
\begin{aligned}
&f_{\theta_{sub}} = Sampling(f_{\theta}) \\
&\mu_{x} = D[p(y_{pred}|x; \theta_{sub}^1), ... , p(y_{pred}|x; \theta_{sub}^n)]
\end{aligned}
\end{equation}
where $D$ computes the statistical discrepancy of $n$ outputs. $\mu_{x}$ is the pixel/voxel- level uncertainty. In the original MC-Dropout model, the dropout operation samples $n$ sub-models $f_{\theta_{sub}}$ in $n$ forward passes.

To address the issue that the MC-Dropout algorithm requires a lot of forward passes (more than eight times in \citep{uamt,pseudolabeling}), we design our proposed MC-Net+ model with one encoder and $n$ slightly different decoders, as shown in Fig.~\ref{model}. With a shared encoder $f_{\theta_e}$, we pre-define $n$ sub-models before the uncertainty estimation. In this way, the uncertainty $\mu_{x}$ of an input $x$ becomes:
\begin{equation}
\begin{aligned}
&f_{\theta_{sub}^{i}} = f_{\theta_{e}} \boxplus f_{\theta_{d}^{i}},\ i\in 1,...,n\\
&\mu_{x} = D[p(y_{pred}|x; \theta_{sub}^1), ..., p(y_{pred}|x; \theta_{sub}^n)]
\end{aligned}
\end{equation}
where the symbol $\boxplus$ means that a sub-model $f_{\theta_{sub}^{i}}$ is composed of one shared encoder $f_{\theta_e}$ and one decoder $f_{\theta_d^i}$. Here, each sub-model $f_{\theta_{sub}^i}$ is a standard encoder-decoder architecture like V-Net or U-Net \citep{vnet,unet}.

Specifically, to achieve a nice trade-off between effectiveness and efficiency, $n$ is set as 3 in this paper (see Fig.~\ref{model}). Here, we employ the transposed convolutional layer, the linear interpolation layer, and the nearest interpolation layer to construct three slightly different sub-models, aiming at increasing the intra-model diversity. In this way, we can approximate the model uncertainty more accurately and then achieve better performance of semi-supervised medical image segmentation. These settings are further discussed in Section \textit{6.2}.

\subsection{Training via Mutual Consistency Constraints}
Based on such a model design, the discrepancy of multiple model outputs is used to represent the model's uncertainty, which indicates the challenging regions. Then, considering that both the consistency and entropy-minimization constraints enable the model to exploit unlabeled data effectively, we propose a novel mutual consistency training strategy, applying two aforementioned constraints to train our model.

Specifically, using a sharpening function \citep{sharpening}, we first convert an output probability map $p(y_{pred}|x; \theta)$ into a soft pseudo label $p^*(y_{pred}^*|x; \theta)$ defined as:
\begin{equation}
p^*(y_{pred}^*|x; \theta) = \frac{p(y_{pred}|x; \theta)^{1/T}}{p(y_{pred}|x; \theta)^{1/T}+(1-p(y_{pred}|x; \theta))^{1/T}}
\end{equation}
where $T$ is a hyper-parameter to control the temperature of sharpening. Appropriate $T$ not only can enforce the entropy minimization constraint to regularize our model, but also would not introduce more noises and confuse the model training. We refer the readers to Section \textit{6.3} for the discussion.

Then, we perform the mutual learning \citep{mutual} between one decoder's probability output and other decoders' soft pseudo labels. In this way, the discrepancy of $n$ outputs is reduced to guide the model learning and the predictions in these highly uncertain regions should be consistent.
The advantages of such a design can be concluded as: (1) the consistency constraints are enforced via encouraging the invariant outputs of all sub-models; (2) under the supervision of soft pseudo labels, the model is learned to generate low-entropy results as the entropy-minimization constraint; (3) the MC-Net+ model can be trained in an `end-to-end' manner without multiple forward passes.

Finally, we employ a weighted sum of a supervised loss and a mutual consistency loss to train our proposed MC-Net+ model as the following:
\begin{align}
&L_{mc}= \sum_{i,j=1 \ \& \ i\neq j}^{n} D[p^*(y_{pred}^*|x; \theta_{sub}^i), p(y_{pred}|x; \theta_{sub}^j)]\\
&Loss =  \lambda\times \sum_{i=1}^{n}L_{seg}(p(y_{pred}|x_l;\theta_{sub}^i),y_l) + \beta\times L_{mc}
\end{align}
where $L_{seg}$ is the popular Dice loss for the segmentation task, and $D$ is the Mean Squared Error (MSE) loss with paired inputs, \textit{i.e.,} $p^*(y_{pred}^*|x; \theta_{sub}^i)$ and $p(y_{pred}|x; \theta_{sub}^j)$. $\lambda$ and $\beta$ are two hyper-parameters to balance the supervised loss $L_{seg}$ and the mutual consistency loss $L_{mc}$. Note that, the $L_{mc}$ is applied on both labeled and unlabeled sets $\mathbb{D_L}$ and $\mathbb{D_U}$.

\section{Experiment}
\subsection{Datasets}
We evaluated the proposed MC-Net+ model on the LA, Pancreas-CT and ACDC datasets. The LA dataset \citep{ladata},  the benchmark dataset for the 2018 Atrial Segmentation Challenge\footnote{\url{http://atriaseg2018.cardiacatlas.org}}, contains 100 gadolinium-enhanced MR imaging scans for training, with an isotropic resolution of $0.625\times0.625\times0.625$ mm. Since the testing set on LA does not include public annotations, following existing models \citep{uamt,sassnet,dtc}, we used a fixed split\footnote{\url{https://github.com/yulequan/UA-MT/tree/master/data}} that 80 samples are training and the rest 20 samples are for validation. Then, we report the performance of our model and other methods on the same validation set for fair comparisons.

The Pancreas-CT dataset \citep{pancreasct} contains 82 3D abdominal contrast-enhanced CT scans, which were collected from 53 male and 27 female subjects at the National Institutes of Health Clinical Center\footnote{\url{https://wiki.cancerimagingarchive.net/display/Public/Pancreas-CT}}. These slices are collected on Philips and Siemens MDCT scanners and have a fixed resolution of $512\times512$ with varying thicknesses from 1.5 to 2.5 mm. The data split is fixed in this paper as the DTC model \citep{dtc}. We employed 62 samples for training and reported the performance of the rest 20 samples. We here clipped the voxel values to the range of $[-125, 275]$ Hounsfield Units (HU) as \citep{ctprocess} and further re-sampled the data into an isotropic resolution of $1.0\times1.0\times1.0$ mm.

The ACDC (Automated Cardiac Diagnosis Challenge) dataset was collected from real clinical exams acquired at the University Hospital of Dijon\footnote{\url{https://www.creatis.insa-lyon.fr/Challenge/acdc/databases.html}} \citep{acdc}. The ACDC dataset contains cardiac MR imaging samples (multi-slice 2-D cine MRI) from 100 patients for training. Also, following \citep{benchmark}, we used a fixed data split\footnote{\url{https://github.com/HiLab-git/SSL4MIS/tree/master/data/ACDC}} in the patient level for our experiments, where the new training, validation and testing sets respectively contain 70, 10 and 20 patients' data. Unlike the task is 3D binary segmentation on the LA and Pancreas-CT datasets, we extend our model to the 2D multi-class segmentation on the ACDC dataset. The 2D MC-Net+ model is designed to segment three targets including the myocardium, left and right ventricles from these 2D MR slices.

\subsection{Implementing Details}
\textbf{3D Segmentation:}
Following \citep{uamt,sassnet,dtc}, we first cropped the 3D samples according to the ground truth, with enlarged margins \textit{i.e.} $[10\sim20, 10\sim20, 5\sim10]$ or $[25, 25, 0]$ voxels on LA or Pancreas-CT, respectively. Then, these scans were normalized as zero mean and unit variance. For training, we randomly extracted 3D patches of size $112\times112\times80$ on LA or $96\times96\times96$ on Pancreas-CT.

Afterward, we applied the 2D rotation and flip operations on the LA dataset as data augmentation. Then, on both datasets, the batch size was set as 4 and each batch contained two labeled patches and two unlabeled patches. The 3D backbone was set as V-Net using the tri-linear interpolation layer to enlarge the feature maps. We trained our 3D MC-Net+ model for 15k iterations. For testing, we employed a sliding window of size $112\times112\times80$ or $96\times96\times96$ with a fixed stride $18\times18\times4$ or $16\times16\times16$ to extract patches on LA or Pancreas-CT, respectively. Then, we recomposed the patch-based predictions as final entire results.

\textbf{2D Segmentation:}
On the ACDC dataset, we also normalized the samples as zero mean and unit variance. The random rotation and flip operations were used to augment data. The 2D patches of size $256\times256$ were randomly extracted and the batch size was set as 24. Each batch included 12 labeled data and 12 unlabeled samples. In the testing time, we resized the scans to $256\times256$ as inputs and then enlarged it to the original size as final results. Our 2D MC-Net+ adopted the U-Net model as the backbone, which utilizes the bi-linear interpolation to expand the feature maps. The 2D model was trained via 30k iterations. All settings on the ACDC dataset followed the public benchmark \citep{benchmark} for fair comparisons.

On all datasets, we adopted the SGD optimizer with a learning rate $10^{-2}$ and a weight decay factor $10^{-4}$ for training. $T$ was set as 0.1. $\lambda$ was 1 for 2D segmentation and 0.5 for 3D tasks. The weight $\beta$ was set as a time-dependent Gaussian warming-up function \citep{rampup} as public methods \citep{uamt,sassnet,dtc,mcnet}. Note that, we performed two typical semi-supervised experimental settings \textit{i.e.,} training with 10\% or 20\% labeled data and the rest unlabeled data, as \citep{uamt,sassnet,dtc}. We re-implemented all compared methods and conducted the experiments in an identical environment (Hardware: Intel(R) Xeon(R) Gold 6150 CPU@2.70GHz, NVIDIA Tesla V100 GPU; Software: PyTorch 1.8.0, CUDA 11.2 and Python 3.8.10; Random Seed: 1337). Following \citep{uamt,sassnet,dtc,mcnet}, we adopted four metrics including Dice, Jaccard, the average surface distance (ASD) and the 95\% Hausdorff Distance (95HD) for the quantitative evaluation.

\section{Result}
\begin{figure*}[htb]
\centering
\includegraphics[width=1\textwidth]{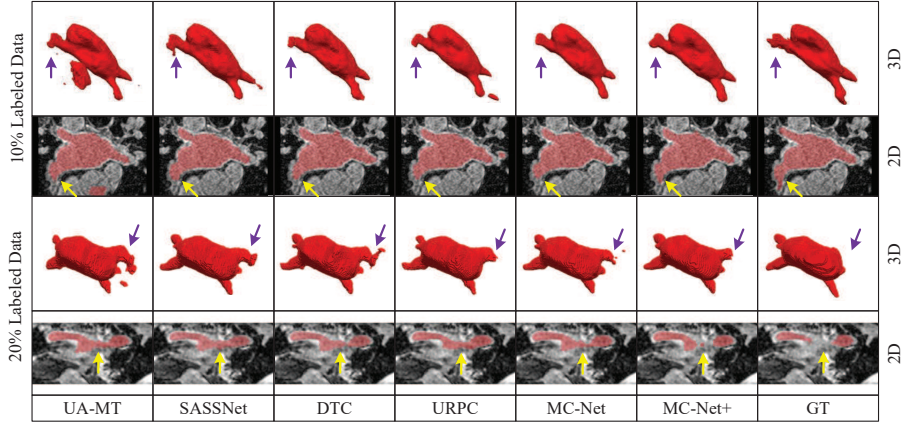}
\caption{\label{result_la}From left to right, there are several exemplar results in 2D and 3D views obtained by UA-MT \citep{uamt}, SASSNet \citep{sassnet}, DTC \citep{dtc}, URPC \citep{dtc}, MC-Net \citep{mcnet}, and our MC-Net+ model on the LA dataset, equipped with the corresponding ground truth (right).}
\end{figure*}
{
\begin{table*}[!htb]
	\centering
	\caption{Comparisons with five state-of-the-art methods on the LA dataset. Note that, the model complexities, \textit{i.e., the number of parameters (Para.) and multiply-accumulate operations (MACs),} are measured during the model inference.}
	\label{tabla}
    \begin{threeparttable}
	\resizebox{\textwidth}{!}{
	\begin{tabular}{c|cc|cccc|cc}
		\hline 
		\hline
		\multirow{2}{*}{Method}&\multicolumn{2}{c}{\# Scans used}&\multicolumn{4}{|c}{Metrics}&\multicolumn{2}{|c}{Complexity}\\
		\cline{2-9}
		&Labeled&Unlabeled &Dice(\%)$\uparrow$ &Jaccard(\%)$\uparrow$&95HD(voxel)$\downarrow$&ASD(voxel)$\downarrow$&Para.(M)&MACs(G)\\
		\hline
		V-Net & 8(10\%) &0 &78.57 &66.96 &21.20 &6.07 &9.44 &47.02\\
		V-Net &16(20\%) &0 &86.96 &77.31 &11.85 &3.22 &9.44 &47.02\\
		V-Net &80(All) &0 &91.62 &84.60 &5.40 &1.64 &9.44 &47.02\\
		\hline
		UA-MT \citep{uamt} (MICCAI) & \multirow{6}{*}{8 (10\%)} &\multirow{6}{*}{72 (90\%)} &86.28 &76.11 &18.71 &4.63 &9.44 &47.02\\
		SASSNet \citep{sassnet} (MICCAI) &  & &85.22 &75.09 &11.18 &2.89 &9.44 &47.05\\
		DTC \citep{dtc} (AAAI) &  & &87.51 &78.17 &8.23 &2.36 &9.44 &47.05\\
		URPC \citep{urpc} (MICCAI) &  & &85.01 &74.36 &15.37 &3.96 &5.88&69.43\\
		MC-Net \citep{mcnet} (MICCAI) &  & &87.50 &77.98 &11.28 & 2.30 &12.35 &95.15\\
		MC-Net+ (Ours) &  & &\textbf{88.96} &\textbf{80.25} &\textbf{7.93} &\textbf{1.86}&9.44 &47.02\\
		\hline
		UA-MT \citep{uamt} (MICCAI) & \multirow{6}{*}{16 (20\%)} &\multirow{6}{*}{64 (80\%)} &88.74 &79.94 &8.39 &2.32 &9.44 &47.02\\
		SASSNet \citep{sassnet} (MICCAI) &  & &89.16 &80.60 &8.95 &2.26 &9.44 &47.05\\
		DTC \citep{dtc} (AAAI)&  & &89.52 &81.22 &7.07 &1.96 &9.44 &47.05\\
		URPC \citep{urpc} (MICCAI)  &  & &88.74 &79.93 &12.73 &3.66 &5.88&69.43\\
		MC-Net \citep{mcnet} (MICCAI) &  & &90.12 &82.12 &8.07 &1.99 &12.35 &95.15\\
		MC-Net+ (Ours) &  & &\textbf{91.07} &\textbf{83.67} &\textbf{5.84} &\textbf{1.67}&9.44 &47.02\\
		\hline
		\hline
	\end{tabular}}
    \end{threeparttable}
\end{table*}
}
\subsection{Performance on the LA Dataset}
Fig.~\ref{result_la} gives several segmentation results of two samples in both 2D and 3D views on the LA dataset. They are obtained by five recent models and our method from left to right. It can be seen that the MC-Net+ model generates a more complete left atrium than other SOTA methods. Note that, we do not use any morphological operations to refine the segmented results \textit{e.g.} selecting the largest connected component as the post-processing module \citep{sassnet}. Our model naturally eliminates most of isolated regions and preserves more fine details (indicated by purple and yellow arrows in Fig.~\ref{result_la}) for the semi-supervised left atrium segmentation.

Table~\ref{tabla} gives the quantitative results on the LA dataset. It also shows the results of a fully supervised V-Net model trained with 10\%, 20\% and all labeled data as the reference. By effectively leveraging the unlabeled data, our proposed MC-Net+ model achieves impressive performance gains from 55\% to 70\% of Dice with only 10\% labeled data training. Meanwhile, the model with only 20\% labeled data training obtains comparable results \textit{e.g.,} 91.07\% vs. 91.62\% of Dice, comparing with the upper bound (V-Net with 100\% labeled data training). At the same time, as depicted in Table~\ref{tabla},  our MC-Net+ model significantly outperforms the other methods in two semi-supervised settings and does not introduce more inference costs compared to the V-Net backbone. 

\begin{figure*}[htb]
\centering
\includegraphics[width=1\textwidth]{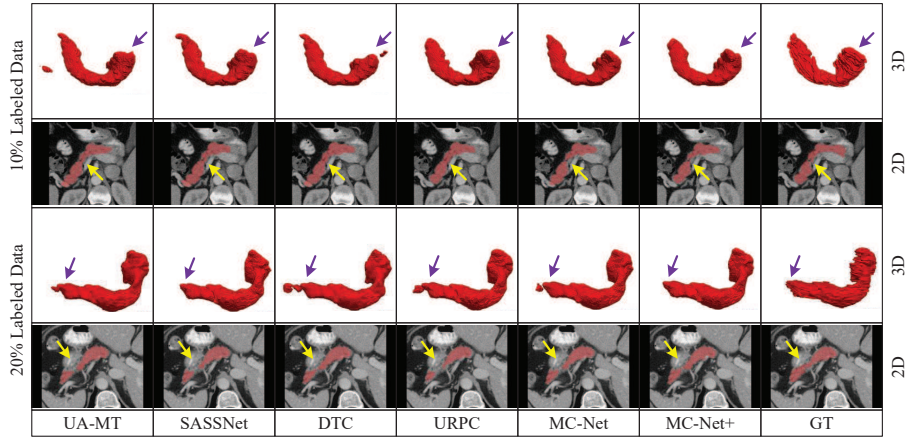}
\caption{\label{result_pa}From left to right, there are several exemplar results in 2D and 3D views obtained by UA-MT \citep{uamt}, SASSNet \citep{sassnet}, DTC \citep{dtc}, URPC \citep{dtc}, MC-Net \citep{mcnet}, and our MC-Net+ model on the Pancreas-CT dataset, equipped with the corresponding ground truth (right).}
\end{figure*}
{
\begin{table*}[!htb]
	\centering
	\caption{Comparisons with five state-of-the-art methods on the Pancreas-CT dataset. Note that, the model complexities, \textit{i.e., the number of parameters (Para.) and multiply-accumulate operations (MACs),} are measured during the model inference.}
	\label{tabpa}
    \begin{threeparttable}
	\resizebox{\textwidth}{!}{
	\begin{tabular}{c|cc|cccc|cc}
		\hline 
		\hline
		\multirow{2}{*}{Method}&\multicolumn{2}{c}{\# Scans used}&\multicolumn{4}{|c}{Metrics}&\multicolumn{2}{|c}{Complexity}\\
		\cline{2-9}
		&Labeled&Unlabeled &Dice(\%)$\uparrow$ &Jaccard(\%)$\uparrow$&95HD(voxel)$\downarrow$&ASD(voxel)$\downarrow$&Para.(M)&MACs(G)\\
		\hline
		V-Net &6 (10\%) &0 &54.94 &40.87 &47.48 &17.43 &9.44&41.45\\
		V-Net  & 12 (20\%) &0 &71.52 &57.68 &18.12 &5.41 &9.44&41.45 \\
		V-Net  &62 (All) &0 &82.60 &70.81 &5.61 &1.33 &9.44&41.45\\
		\hline
		UA-MT \citep{uamt} (MICCAI)& \multirow{7}{*}{6 (10\%)} &\multirow{7}{*}{56 (90\%)} &66.44 &52.02 &17.04 &3.03 &9.44&41.45\\
		SASSNet \citep{sassnet}  (MICCAI)  &  & &68.97 &54.29 &18.83 &\textbf{1.96}&9.44 &41.48\\
		DTC \citep{dtc}  (AAAI)  &  & &66.58 &51.79 &15.46 &4.16 &9.44 &41.48\\
		URPC \citep{urpc} (MICCAI)  &  & &\textbf{73.53} &\textbf{59.44} &22.57 &7.85 &5.88 &61.21\\
		MC-Net \citep{mcnet}  (MICCAI)  &  & &69.07 &54.36 &\textbf{14.53} &2.28 &12.35&83.88\\
		MC-Net+  (Ours) & & &70.00 &55.66 &16.03 &3.87&9.44&41.45\\
		\textit{Multi-scale MC-Net+$^*$} & & &\textbf{\textit{74.01}} &\textbf{\textit{60.02}} &\textbf{\textit{12.59}} &\textit{3.34} &5.88 &61.21\\
		\hline
		UA-MT \citep{uamt}  (MICCAI) & \multirow{7}{*}{12 (20\%)} &\multirow{7}{*}{50 (80\%)} &76.10 &62.62 &10.84 &2.43 &9.44&41.45\\
		SASSNet \citep{sassnet}  (MICCAI)  &  & &76.39 &63.17 &11.06 &\textbf{1.42}&9.44 &41.48\\
		DTC \citep{dtc} (AAAI)  &  & &76.27 &62.82 &8.70 &2.20 &9.44 &41.48\\
		URPC \citep{urpc} (MICCAI)  &  & &\textbf{80.02} &\textbf{67.30} &8.51 &1.98 &5.88 &61.21\\
		MC-Net \citep{mcnet}  (MICCAI)  &  & &78.17 &65.22 &\textbf{6.90} &1.55 &12.35&83.88\\
		MC-Net+  (Ours)  &  & &79.37 &66.83 &8.52 &1.72&9.44&41.45\\
		\textit{Multi-scale MC-Net+$^*$} & & &\textbf{\textit{80.59}} &\textbf{\textit{68.08}} &\textbf{\textit{6.47}} &\textit{1.74} &5.88 &61.21\\
		\hline
		\hline
	\end{tabular}}
    \begin{tablenotes}
    \footnotesize
    \item[*]We designed our multi-scale MC-Net+ model based on \citep{urpc}.
    \end{tablenotes}
    \end{threeparttable}
\end{table*}
}

\subsection{Performance on the Pancreas-CT Dataset}
Fig.~\ref{result_pa} and Table~\ref{tabpa} show the corresponding results of our model and five semi-supervised methods on the Pancreas-CT dataset. Except for the multi-scale consistency method \citep{urpc}, our proposed MC-Net+ model achieved the highest Dice and Jaccard than other methods for semi-supervised pancreas segmentation. Here, the original mutual consistency constraint is only performed at the single scale. However, the pancreas segmentation is a relatively difficult task and may require more multi-scale information. Therefore, based on \citep{urpc}, we further design a new multi-scale MC-Net+ model, achieving the best performance in each setting on the Pancreas-CT dataset, see Table~\ref{tabpa}. It demonstrates that our proposed model can be easily incorporated with other multi-scale methods to further improve the segmentation performance. Moreover, our model does not rely on any post-processing modules and we do not use any shape-related constraints to train our model. Similar with the results on the LA dataset, our single-scale MC-Net+ model is able to obtain comparable performance in terms of the surface-based metrics and can accurately segment the challenging areas, indicated by the purple and yellow arrows in Fig.~\ref{result_pa}, on the Pancreas-CT dataset.

{
\begin{table*}[!htb]
	\centering
	\caption{Comparisons with five state-of-the-art methods on the ACDC dataset. Note that, the model complexities, \textit{i.e., the number of parameters (Para.) and multiply-accumulate operations (MACs),} are measured during the model inference.}
	\label{tabacdc}
    \begin{threeparttable}
	\resizebox{\textwidth}{!}{
	\begin{tabular}{c|cc|cccc|cc}
		\hline 
		\hline
		\multirow{2}{*}{Method}&\multicolumn{2}{c}{\# Scans used}&\multicolumn{4}{|c}{Metrics}&\multicolumn{2}{|c}{Complexity}\\
		\cline{2-9}
		&Labeled&Unlabeled &Dice(\%)$\uparrow$ &Jaccard(\%)$\uparrow$&95HD(voxel)$\downarrow$&ASD(voxel)$\downarrow$&Para.(M)&MACs(G)\\
		\hline
		U-Net & 7 (10\%) &0 &77.34 &66.20 &9.18 &2.45 &1.81&2.99\\
		U-Net & 14 (20\%) &0 &85.15 &75.48 &6.20 &2.12 &1.81&2.99\\
		U-Net & 70 (All) &0 &91.65 &84.93 &1.89 &0.56 &1.81&2.99\\
		\hline
		UA-MT \citep{uamt} (MICCAI)& \multirow{6}{*}{7 (10\%)} &\multirow{6}{*}{63 (90\%)} &81.58 &70.48 &12.35 &3.62 &1.81&2.99\\
		SASSNet \citep{sassnet} (MICCAI)  &  & &84.14 &74.09 &\textbf{5.03} &\textbf{1.40}&1.81&3.02\\
		DTC \citep{dtc} (AAAI)  &  & &82.71 &72.14 &11.31 &2.99 &1.81&3.02\\
		URPC \citep{urpc} (MICCAI) &  & &81.77 &70.85 &5.04 &1.41 &1.83&3.02\\
		MC-Net \citep{mcnet} (MICCAI)  &  & &86.34 &76.82 &7.08 &2.08&2.58&5.39\\
		MC-Net+ (Ours) & & &\textbf{87.10} &\textbf{78.06} &6.68 &2.00&1.81&2.99\\
		\hline
		UA-MT \citep{uamt} (MICCAI) & \multirow{6}{*}{14 (20\%)} &\multirow{6}{*}{56 (80\%)} &85.87 &76.78 &5.06 &1.54 &1.81&2.99\\
		SASSNet \citep{sassnet} (MICCAI)  &  & &87.04 &78.13 &7.84 &2.15&1.81&3.02\\
		DTC \citep{dtc} (AAAI)  &  & &86.28 &77.03 &6.14 &2.11 &1.81&3.02\\
		URPC \citep{urpc} (MICCAI) &  & &85.07 &75.61 &6.26 &1.77 &1.83&3.02\\
		MC-Net \citep{mcnet} (MICCAI)  &  & &87.83 &79.14 &\textbf{4.94} &\textbf{1.52}&2.58&5.39\\
		MC-Net+ (Ours)  &  & &\textbf{88.51} &\textbf{80.19} &5.35 &1.54&1.81&2.99\\
		\hline
		\hline
	\end{tabular}}
    \end{threeparttable}
\end{table*}
}
\begin{figure*}[htb]
\centering
\includegraphics[width=1\textwidth]{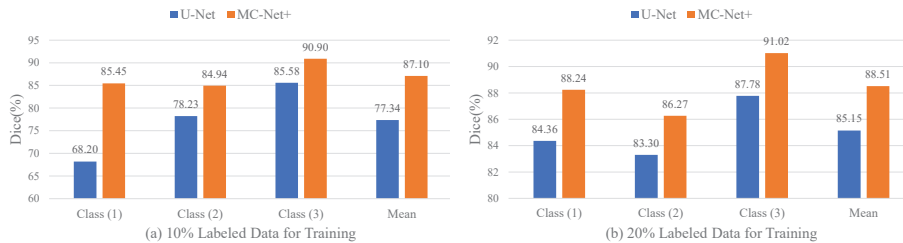}
\caption{\label{result_acdc}Dice performance of each class on the ACDC dataset, obtained by a fully supervised U-Net model and our semi-supervised MC-Net+ model, with 10\% \textit{(a)} and 20\% \textit{(b)} labeled data for training.}
\end{figure*}
\subsection{Performance on the ACDC Dataset}
We further extended our model for the 2D multi-class segmentation task. The results in Table~\ref{tabacdc} are the average performance of three segmented targets, \textit{i.e., the myocardium, left and right ventricles} on the ACDC dataset. It indicates that: (1) comparing with other methods, our model obtained the highest Dice, Jaccard and comparable surface-related performance in each semi-supervised setting; (2) via exploiting the unlabeled data effectively, our model almost produced an average Dice gain of 10\% or 3\% than the fully-supervised U-Net model trained with 10\% or 20\% labeled data. Additionally, Fig.~\ref{result_acdc} shows the dice performance of each class, obtained by the fully-supervised U-Net model and our semi-supervised MC-Net+ model. We can see that, either with 10\% or 20\% labeled data for training, our MC-Net+ model achieves impressive performance gains in each class for 2D medical image segmentation.

Overall, based on the results on three datasets, our MC-Net+ model shows superior performance than other SOTA methods for semi-supervised medical image segmentation. Note that, our model does not rely on specific backbones and can be applied for various medical tasks in either 2D or 3D segmentation. Meanwhile, it demonstrates that adding another decoder to increase the model's diversity leads to the improved semi-supervised segmentation performance on all datasets (\textit{i.e.,} MC-Net+ model vs. MC-Net model \citep{mcnet}).

Furthermore, we observe three interesting properties of our proposed model: (1) the model without any shape-related constraints can generate satisfied segmented results with fewer isolated regions; (2) our MC-Net+ is effective in segmenting some challenging regions \textit{e.g.,} thin branch joints in Fig.~\ref{result_la}; (3) for testing, the MC-Net+ model does not introduce additional inference costs. These properties are beneficial for constructing an automatic CAD system to diagnose relevant diseases in practical clinical analysis.

\section{Discussion}
\subsection{Ablation Study}
{
\begin{table*}[!htb]
	\centering
	\caption{Ablation studies of our MC-Net+ model on the LA dataset. Note that, \textit{DD} means using different up-sampling strategies to construct three decoders, \textit{CC} means only enforcing the consistency constraints for only $p(y_{pred}|x; \theta_{sub})$ or $p^*(y_{pred}^*|x; \theta_{sub})$, and \textit{MC} means applying the mutual consistency constraints between $p(y_{pred}|x; \theta_{sub})$ and $p^*(y_{pred}^*|x; \theta_{sub})$.}
	\label{tabablation}
    \begin{threeparttable}
	\resizebox{\textwidth}{!}{
	\begin{tabular}{cc|ccc|cccc}
		\hline 
		\hline
		\multicolumn{2}{c|}{\# Scans used}&\multicolumn{3}{c|}{Designs}&\multicolumn{4}{c}{Metrics}\\
		\hline
		Labeled&Unlabeled &\textit{DD}&\textit{CC}&\textit{MC}&
		Dice(\%)$\uparrow$ &Jaccard(\%)$\uparrow$&95HD(voxel)$\downarrow$&ASD(voxel)$\downarrow$\\
		\hline
		8 (10\%)&0& &&& 82.69&70.93&28.27 &7.89 \\
		\hline
		\multirow{6}{*}{8 (10\%)} &\multirow{6}{*}{72 (90\%)}& & \checkmark&&87.61&78.17&13.65 &3.09 \\
		&& &$\checkmark^{*}$& &88.33 &79.31 &9.17 &1.92 \\
		&& &&\checkmark&88.58 &79.68&\textbf{7.79} &2.01 \\
		&& \checkmark&\checkmark&&88.65 &79.77&9.24 &2.20 \\
		&& \checkmark&$\checkmark^{*}$&&88.70 &79.85&8.08 &2.03 \\
		&& \checkmark& &\checkmark&\textbf{88.96} &\textbf{80.25} &7.93 &\textbf{1.86} \\
		\hline
		16 (20\%)&0& &&&86.01 &75.92&19.27 &4.84 \\
		\hline
		\multirow{6}{*}{16 (20\%)} &\multirow{6}{*}{64 (80\%)}& & \checkmark&&90.60&82.90&7.44&2.28 \\
		&& &$\checkmark^{*}$&&90.60 &82.91&6.13&1.88 \\
		&& &&\checkmark& 90.84&83.32&5.89 &1.85 \\
		&& \checkmark&\checkmark&&90.77 &83.20&8.27 &2.50 \\
		&& \checkmark&$\checkmark^{*}$&&90.63 &83.03&5.99 &\textbf{1.61} \\
		&& \checkmark&&\checkmark &\textbf{91.07} &\textbf{83.67} &\textbf{5.84} &1.67 \\
		\hline
		80 (100\%)&0& &&&92.05 &85.33&7.10 &1.79 \\
		\hline
		\hline
	\end{tabular}}
    \begin{tablenotes}
    \footnotesize
    \item[*]The consistency constraints are enforced for $p^*(y_{pred}^*|x; \theta_{sub})$ of different sub-models.
    \end{tablenotes}
    \end{threeparttable}
\end{table*}
}
The ablation studies (see Table~\ref{tabablation}) were conducted on the LA dataset, to show the effectiveness of each design. It reveals that, trained with 10\% or 20\% labeled data, (1) the most significant performance gains (\textit{the average dice gains are 5.28\% and 4.59\%, respectively}) are achieved by forcing three decoders to generate similar results (\textit{i.e.,} reducing the model uncertainty); (2) using multiple slightly different decoders, labeled by \textit{DD}, results in average dice gains of 0.63\% and 0.13\%. Note that, a concurrent work \citep{cps} used identical model architectures with different initialization parameters while we employ different up-sampling strategies to further increase the intra-model diversity, leading to a better performance; (3) encouraging the mutual consistency for training, labeled by \textit{MC}, is always better than applying consistency constraints for probability outputs or soft pseudo labels, labeled by \textit{CC} or $\textit{CC}^*$. 
We also provide the fully supervised MC-Net+ model\textit{, i.e.,} without $L_{mc}$ for training, as the reference. The results show that simply adopting three slightly different decoders does not bring impressive performance gains while enforcing our novel mutual consistency constraints can significantly improve the semi-supervised segmentation performance on the LA dataset, with overall dice gains of 6.25\% and 5.07\% under both settings, respectively.

\subsection{Effects of Different Up-sampling Strategies}
\begin{figure*}[htb]
\centering
\includegraphics[width=1\textwidth]{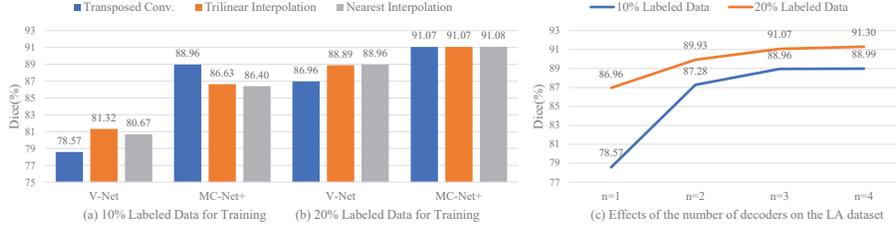}
\caption{\label{dicussion} Dice performance of different up-sampling strategies \textit{(a} and \textit{b)} and the number of decoders \textit{(c)} on the LA dataset. Note that, to construct four decoders, we use two transposed convolutional layers with different initialization parameters for training due to the limitation of public available up-sampling strategies.}
\end{figure*}
To increase the intra-model diversity, our MC-Net+ model adopts the transposed convolutional layer, the linear interpolation layer, and the nearest interpolation layer to construct three slightly different decoders. Fig.~\ref{dicussion} indicates that, the dice values of fully supervised V-Net models with different up-sampling strategies vary significantly on the LA dataset. However, when applying our mutual consistency constraints, our MC-Net+ model obtains better results and three slightly decoders tend to generate invariant outputs, leading to fewer ambiguous predictions and lower model uncertainty. Since three decoders can generate similar results, we only select the original encoder-decoder architecture\textit{, i.e.,} the shared encoder and the first decoder, as the final testing model to reduce the inference costs. Note that, our previous MC-Net model \citep{mcnet} employs the mean results of two decoders as final predictions while the new MC-Net+ model only uses the first output in the experiments.

Furthermore, since the number $n$ of decoders is scalable, we further conducted a sensitivity experiment to show the effects of $n$. Fig.~\ref{dicussion} (c) shows that introducing more decoders can improve the performance, but the gains are decreased due to the confirmation bias \citep{dividemix}. In other words, since the labeled data is extremely limited, deep models may generate wrong predictions but with high confidence. Therefore, $n$ is set as 3 in this paper to achieve a balance between effectiveness and efficiency. We also believe that if the labeled data is adequate, our model with more diverse sub-models can perform better in medical image segmentation.

\subsection{\texorpdfstring{Effects of Temperature $T$}{}}
\begin{figure*}[htb]
\centering
\includegraphics[width=0.9\textwidth]{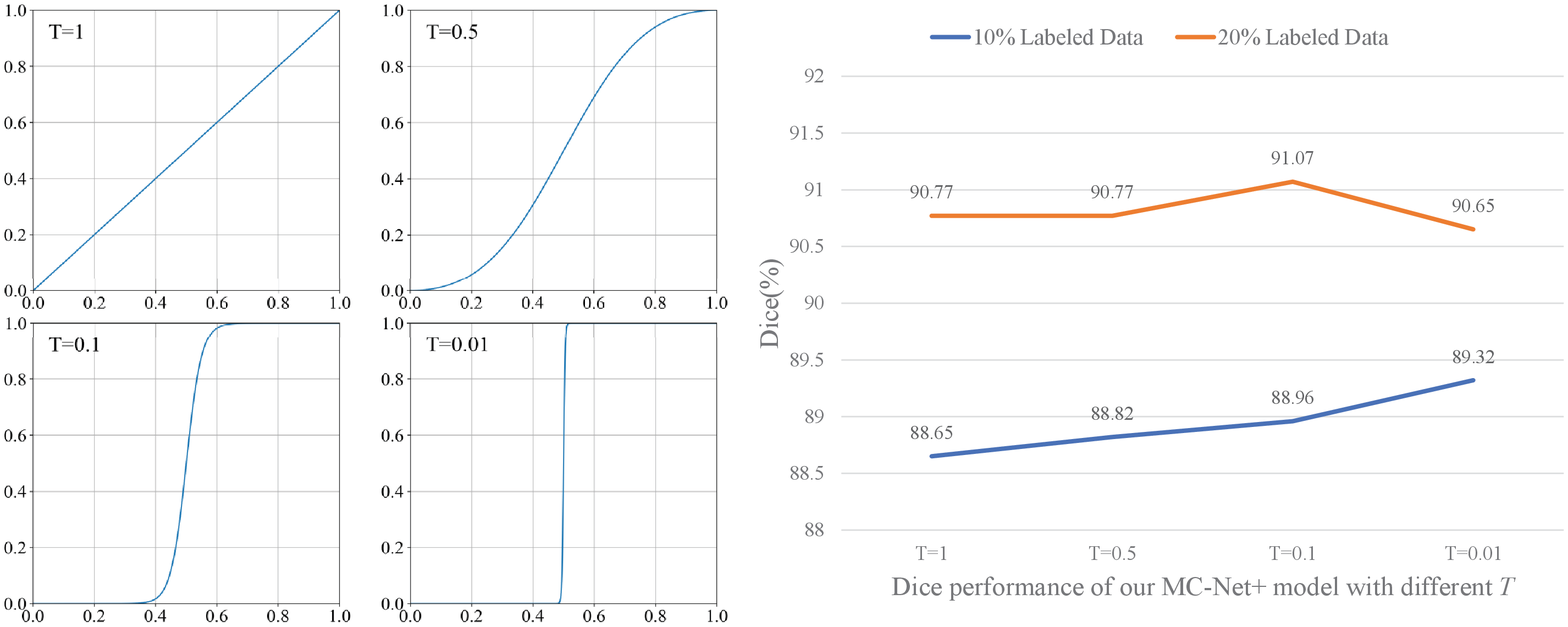}
\caption{\label{temperature} Illustrations of corresponding sharpening functions \textit{(left)} and dice performance \textit{(right)} with different sharpening temperatures $T$ on the LA dataset.}
\end{figure*}
To mitigate the effects of easily mis-classified pixels, the MC-Net+ model uses soft pseudo labels to apply the entropy-minimization constraint and does not significantly sharpen the plausible predictions around probability 0.5 (see the \textit{left} of Fig.~\ref{temperature}). Fig.~\ref{temperature} \textit{right} gives the dice performance of our MC-Net+ model trained with different temperatures $T$ on the LA dataset. It shows that, in each semi-supervised setting, the dice values of different $T$ are similar, which indicates that our model is relatively robust about the hyper-parameter $T$. Here, a larger $T$ cannot enforce sufficient entropy-minimization constraints for the model training while a smaller $T$ may increase the noises of pseudo labels, leading to the error acclamation. Therefore, we finally adopt the sharpening function with temperature 0.1 to generate soft pseudo labels on all datasets.

\subsection{\texorpdfstring{Effects of Loss Weight $\lambda$}{}}
\begin{figure*}[htb]
\centering
\includegraphics[width=1\textwidth]{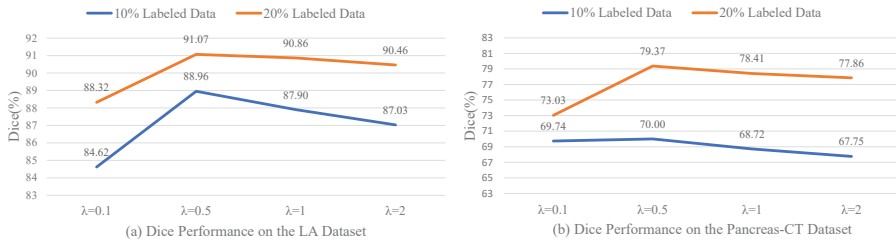}
\caption{\label{weight} Dice performance with different loss weights $\lambda$ on the LA and Pancreas-CT datasets.}
\end{figure*}
We further conducted a parameter sensitivity experiment on the LA and Pancreas-CT datasets, to show the effects of $\lambda$ for the balance of two losses (see Fig.~\ref{weight}). Here, a smaller $\lambda$ would decrease the performance since three decoders may generate inaccurate results due to the insufficient labeled data training, while a larger $\lambda$ can not apply enough mutual consistency constraints and thus also obtains a sub-optimal performance. Therefore, in this paper, we set the weight $\lambda$ as 0.5 to balance two losses on the LA and Pancreas-CT datasets.
\subsection{\texorpdfstring{Effects of Different Distance Measurements $D$}{}}
{
\begin{table*}[!htb]
	\centering
	\caption{Discussion of different distance measurements $D$ on the LA dataset.}
	\label{tabloss}
    \begin{threeparttable}
	\resizebox{\textwidth}{!}{
	\begin{tabular}{c|c|cc|cccc}
		\hline 
		\hline
		\multirow{2}{*}{Method}&\multirow{2}{*}{Output}&\multicolumn{2}{c}{\# Scans used}&\multicolumn{4}{|c}{Metrics}\\
		\cline{3-8}
		&&Labeled&Unlabeled &Dice(\%)$\uparrow$ &Jaccard(\%)$\uparrow$&95HD(voxel)$\downarrow$&ASD(voxel)$\downarrow$\\
		\hline
		\multirow{2}{*}{MC-Net+ w/ KL} &First Decoder& \multirow{2}{*}{8 (10\%)}
		&\multirow{2}{*}{72 (90\%)} &87.86 &78.86 &9.65 &2.27 \\
		&Mean & & &88.04 &79.08 &9.40 &2.32 \\
		\hline
		\multirow{2}{*}{MC-Net+ w/ MSE}&First Decoder  & \multirow{2}{*}{8 (10\%)} &\multirow{2}{*}{72 (90\%)} &\textbf{88.96} &\textbf{80.25} &\textbf{7.93} &\textbf{1.86} \\
		&Mean & & &\textbf{88.94} &\textbf{80.22} &\textbf{8.02} &\textbf{1.83} \\
		\hline
		\hline
		\multirow{2}{*}{MC-Net+ w/ KL}&First Decoder  & \multirow{2}{*}{16 (20\%)} &\multirow{2}{*}{64 (80\%)} &90.93 &83.45 &6.04 &\textbf{1.63} \\
		&Mean & & &90.96 &83.50 &6.08 &\textbf{1.60} \\
		\hline
		\multirow{2}{*}{MC-Net+ w/ MSE}&First Decoder & \multirow{2}{*}{16 (20\%)} &\multirow{2}{*}{64 (80\%)} &\textbf{91.07} &\textbf{83.67} &\textbf{5.84} &1.67 \\
		&Mean & & &\textbf{91.08} &\textbf{83.70} &\textbf{5.90} &\textbf{1.60} \\
		\hline
		\hline
	\end{tabular}}
    \end{threeparttable}
\end{table*}
}
We discussed the effects of using different $D$ to measure the discrepancy of multiple outputs on the LA dataset. In Table~\ref{tabloss}, we give the results of our MC-Net+ model using the Kullback-Leibler (KL) divergence for training. We can see that the KL loss can also improve the segmentation performance in each semi-supervised setting. Nevertheless, the simple MSE loss is sufficient to demonstrate the effectiveness of our model. Therefore, we finally adopt the MSE loss as $D$ in this paper.

\subsection{Limitations and Future Work}
Although our model is simple and powerful for semi-supervised medical image segmentation, the model design still requires multiple pre-defined decoders, and the selection of existing up-sampling strategies is limited. For new tasks, more varied model architectures are worth exploring to increase the intra-model diversity. Meanwhile, in this paper, we only discuss the model-level perturbations\textit{, i.e.,} using different up-sampling strategies, while the data-level perturbations should also be useful. However, some data-agnostic operations like ColorJitter \citep{fixmatch} may not be suitable for medical data. Future work will focus on developing the data-specific perturbation operations and using more large-scale datasets to evaluate the proposed model.
\section{Conclusion}
In this paper, we have presented a novel MC-Net+ model for semi-supervised medical image segmentation. Effectively leveraging the challenging regions plays an important role in the semi-supervised segmentation. The model design with three slightly different decoders is used to indicate highly uncertain areas and a new mutual consistency constraint between the probability outputs and soft pseudo labels establishes an `end-to-end' way to force the model to generate invariant and low-entropy predictions in the hard regions. Extension experiments demonstrate our model has achieved superior performance over five existing models on three medical datasets and the proposed MC-Net+ model sets a new state of the art for semi-supervised medical image segmentation.
\section{Acknowledgments}
This work was supported in part by the Monash FIT Start-up Grant, and in part by the National Natural Science Foundation of China under Grants 62171377, and in part by the Key Research and Development Program of Shaanxi Province under Grant 2022GY-084. We also appreciate the efforts to collect and share the datasets \citep{ladata,pancreasct,acdc} and several public benchmarks \citep{uamt,sassnet,dtc,urpc,benchmark}.

\bibliography{paper}

\end{document}